\pdfoutput=1

\documentclass[11pt]{article}
\usepackage{acl}

\usepackage{times}
\usepackage{latexsym}

\usepackage[T1]{fontenc}

\usepackage[utf8]{inputenc}
\usepackage{kotex}
\usepackage{microtype}
\usepackage{booktabs}
\usepackage{amsmath}
\usepackage{flexisym}
\usepackage{dcolumn}
\usepackage{kotex}
\usepackage{hhline}
\usepackage{verbatim}
\usepackage{multirow}
\usepackage{amssymb}
\usepackage{rotating}
\usepackage{subfigure}
\usepackage{dblfloatfix}
\usepackage{enumitem}
\usepackage{framed}
\usepackage[utf8]{inputenc}
\usepackage[english]{babel}
\usepackage{xcolor}
\usepackage{tabulary}
\usepackage{color}
\usepackage{arydshln}
\usepackage{ulem}
\usepackage{soul,color}
\usepackage{multirow}
\usepackage{multicol}
\usepackage{xcolor}
\usepackage{graphicx}

\newcolumntype{L}[1]{>{\raggedright\let\newline\\\arraybackslash\hspace{0pt}}m{#1}}
\newcolumntype{C}[1]{>{\centering\let\newline\\\arraybackslash\hspace{0pt}}m{#1}}
\newcolumntype{R}[1]{>{\raggedleft\let\newline\\\arraybackslash\hspace{0pt}}m{#1}}

\author{Hwanhee Lee$^{1}$, Cheoneum Park$^{2}$, Seunghyun Yoon$^{3}$, \\
\bf Trung Bui$^{3}$, Franck Dernoncourt$^{3}$, Juae Kim$^{2}$~\and Kyomin Jung$^{1}$ \\
  $^{1}$Dept. of Electrical and Computer Engineering, Seoul National University \\
  $^{2}$AIRS Company, Hyundai Motor Group,
  $^{3}$Adobe Research\\
  \texttt{\{wanted1007,kjung\}@snu.ac.kr},
  \texttt{\{cheoneum.park, juaekim\}@hyundai.com}\\
  \texttt{\{syoon, bui, franck.dernoncourt\}@adobe.com}
  }
 
\title{Factual Error Correction for Abstractive Summaries Using Entity Retrieval}

\begin{document}
\maketitle

\begin{abstract}
Despite the recent advancements in abstractive summarization systems leveraged from large-scale datasets and pre-trained language models, the factual correctness of the summary is still insufficient.
One line of trials to mitigate this problem is to include a post-editing process that can detect and correct factual errors in the summary. In building such a post-editing system, it is strongly required that 1) the process has a high success rate and interpretability and 2) has a fast running time. Previous approaches focus on regeneration of the summary using the autoregressive models, which lack interpretability and require high computing resources. 
In this paper, we propose an efficient factual error correction system RFEC based on entities retrieval post-editing process. 
RFEC first retrieves the evidence sentences from the original document by comparing the sentences with the target summary. This approach greatly reduces the length of text for a system to analyze.
Next, RFEC detects the entity-level errors in the summaries by considering the evidence sentences and substitutes the wrong entities with the accurate entities from the evidence sentences. Experimental results show that our proposed error correction system shows more competitive performance than baseline methods in correcting the factual errors with a much faster speed.
\end{abstract}
\section{Introduction}
\begin{figure}[!t]
\small
\begin{framed}
\textbf{Article:} \textit{Singer-songwriter David Crosby hit a jogger with his car Sunday evening, a spokesman said.} The accident happened in Santa Ynez, California, near where Crosby lives. Crosby was driving at approximately 50 mph when he struck the jogger, according to California Highway Patrol Spokesman Don Clotworthy. The posted speed limit was 55. The jogger suffered multiple fractures, and was airlifted to a hospital in Santa Barbara, Clotworthy said.,...\\\

\textbf{System Summary with Factual Error:} \textit{\textbf{Don Clotworthy} hit a jogger with his car Sunday evening.} The jogger suffered multiple fractures and was airlifted to a hospital. \\\

\textbf{After Correction:}  \textit{\textbf{David Crosby} hit a jogger with his car Sunday evening.} The jogger suffered multiple fractures and was airlifted to a hospital.
\end{framed}
\caption{
An example of generated summary with factual errors and the correct summary after minor modification.
} 
\vspace{-5mm}
\label{fig_intro}
\end{figure}

\begin{figure*}[!h]
\small
\centering
\includegraphics[width=1.9\columnwidth]{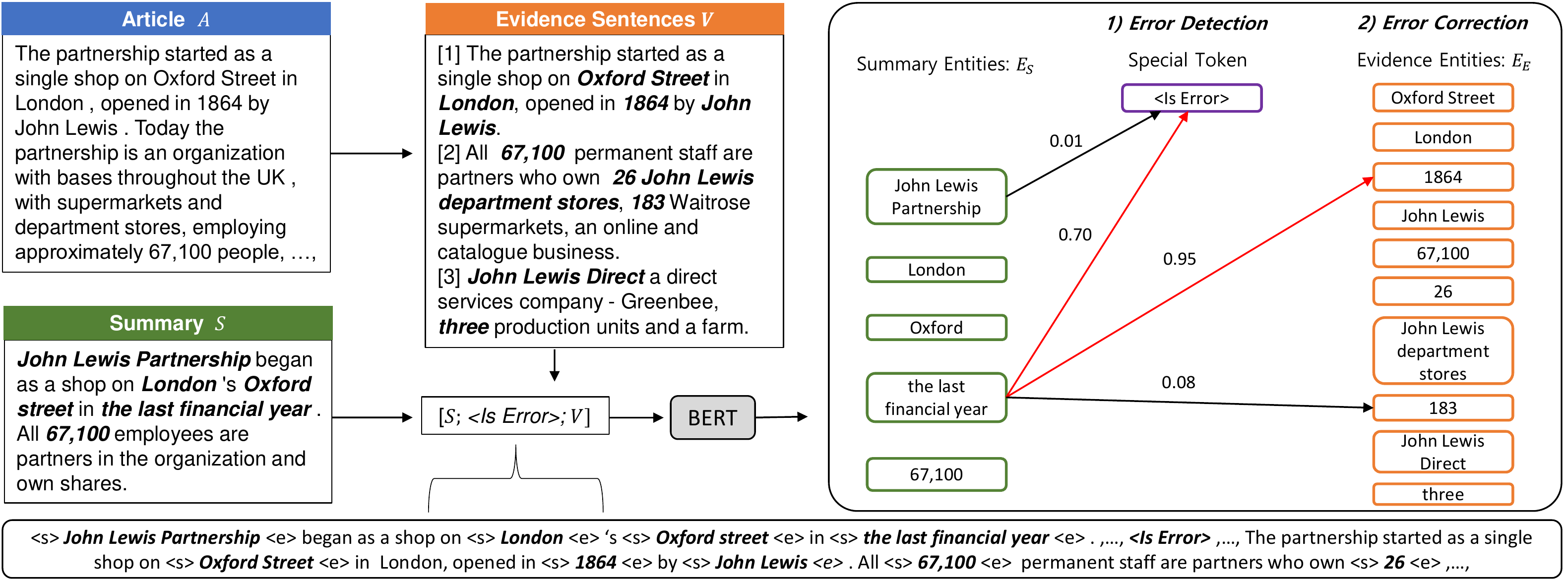}
\caption{
Overall flow of our proposed retrieval-based factual error correction system. Given a summary $S$ and an article $A$, we first retrieve evidence sentences $V$. Using $S$ and $V$, we compute BERT embeddings for entities in summary $E_S$ and evidence sentence $V$. Note that \textit{<Is Error>} is a special token for classifying whether each entity is an error. If the erroneous score computed using \textit{<Is Error>} token is above threshold, we regard those entity as an error and substitute it with one of the entities in the evidence sentences that obtains highest score.
}
\vspace{-5mm}
\label{fig_overall}
\end{figure*}
Text summarization is a task that aims to generate a short version of the text that contains the important information for the given source article. With the advances of neural text summarization systems, abstractive summarization systems~\cite{nallapati2017summarunner} that generate novel sentences rather than extracting the snippets in the source are widely used~\cite{lin2019abstractive}. However, factual inconsistency between the original text and the summary is frequently observed in the abstractive summarization system~\cite{cao2018faithful, zhao-etal-2020-reducing, maynez2020faithfulness} as shown in the system summary of Figure~\ref{fig_intro}. As in the example of Figure~\ref{fig_intro}, many of these errors in the summaries occur at the entry-level such as person name and number.
But these types of errors are sometimes trivial and can often be easily solved through simple modification like changing the wrong entities, as shown in Figure~\ref{fig_intro}. For this reason, previous works~\cite{cao2020factual, zhu2021enhancing} have introduced post-editing systems to alleviate these factual errors in the summary. But all of those works adopt the seq2seq model, which requires a similar cost to the original abstractive summarization systems, as a post-editing. Therefore, using such systems based on seq2seq doubles the inference time for performing post-editing, resulting in significant inefficiency. In addition, seq2seq based post-editing model can be affected by the model's own bias to the input summary.

To overcome this issue and develop efficient factual corrector for summarization systems, we propose a totally different approach, RFEC(\textbf{R}etrieval-based \textbf{F}actual \textbf{E}rror \textbf{C}orrector) that efficiently corrects the factual errors with much faster running time compared to seq2seq model. RFEC first retrieves the evidence sentences for the given summary for correcting and detecting errors. By doing so, we shorten the input length of the model to obtain computational efficiency. Then, RFEC examines all of the entities whether each entity has a factual error. If any entities have a factual error, RFEC substitutes these wrong entities with the correct entity by choosing them among the entities in the source article. Through these steps, we do not create a whole sentence
as in the seq2seq model, but decide whether to fix and correct it through the retrieval, resulting in higher computational efficiency.
Experiments on both synthetic and real-world benchmark datasets demonstrate that our model shows competitive performance with the baseline model with much faster running time. Also, as shown in Figure~\ref{fig_overall}, RFEC has a natural form of interpretability through the visualization of the erroneous score and the scores of each candidate entity for correcting the wrong entities.
\section{Method}
\subsection{Problem Formulation}
For a given summary $S$ and an article $A$, we aim to develop a 
factual error correction system that can fix the possible factual errors in $S$.
Since most of the factual errors appear in entity-level, we develop a system that is specialized in correcting entity-level errors. Specifically, we define this problem as two steps, entity-level error detection and entity-level error correction as shown in Figure~\ref{fig_overall}. For given $ns$ entities $E_{S} = \{es_{1}, es_{2}, ..., es_{ns}\}$ in a summary $S$, we first classify whether each entity is factually consistent with the article $A$. If any entity $e_{S_i}$ is factually inconsistent, the system substitutes it with one of the $na$ entities in the article $E_{A} = \{ea_{1}, ea_{2}, ..., ea_{na} \}$. 

\subsection{Training Dataset Construction}
\label{sssec:sec_trainset}
To train a factual error correction system, we need a triple composed of an input summary $S_{1}$ that may have factual errors, an article $A$ and a target summary $S_{2}$ that is a modified version of $S_{1}$ without factual errors. However, it is difficult to obtain $S_{1}$ that has the errors with the position annotated and the right ground truth correction of such errors. Hence, to train a system, we construct a synthetic dataset by modifying the reference summaries following previous works~\cite{cao2020factual, zhu2021enhancing, kryscinski-etal-2020-evaluating}. We corrupt reference summaries in CNN/DM dataset~\cite{nallapati-etal-2016-abstractive} by randomly changing one of the entities with the same type of other entities in the dataset to make a corrupted summary. Finally, we construct a triple ($S_{1}$, $A$, $S_{2}$). Meanwhile, in the real world dataset, a significant number of summaries are factually consistent, so we only make errors for 50\% of the summaries and set $S_{1}=S_{2}$ for the rest of the summaries in the dataset. Through this procedure, we construct the synthetic training dataset where the number of each train/validation split is 133331/6306, respectively.

\subsection{Evidence Sentence Retrieval}
Generally, a summary does not treat all of the contents in the article but only contains some important parts of the article. Hence, in most cases, checking for errors within the summary and correcting them does not require the entire article, and using the part related to the summary is sufficient, as shown in Figure~\ref{fig_overall}. Inspired by this observation, we extract some of the sentences in the article according to the similarity with the summary to increase the efficiency of the system by shortening the input length. We use ROUGE-L~\cite{lin-2004-rouge} score as a similarity measure to extract top-2 evidence sentences for each sentence in summary. Then, we remove the duplicates and sort them according to the order in which they appear in the article, and combine them to form $V= \{V_{1}, V_{2},...,V_{M}\}$, a set of evidence sentences for detecting and correcting errors in the summary $S$.

\subsection{Entity Retrieval Based Factual Error Correction}
\paragraph{Computing Embedding}
Using summary $S$ and the evidence sentences $V$, we first extract entities $E_{S}$ and $E_{V}$ respectively using SpaCy\footnote{https://spacy.io/api/entityrecognizer} named entity recognition model. And we insert special tokens <s> and <e>, before and after each extracted entity. Then we also insert an additional token \textit{<Is Error>}, which is later used for checking the factual consistency between $S$ and $V$ and concatenate them to make an input for the BERT~\cite{devlin2019bert}. Using BERT, we obtain the contextualized embedding of each entity in $S$ and $V$ as follows:
\begin{equation}
\begin{aligned}
\scriptstyle{H = [{h}_{1},{h}_{2}, ..., {h}_{l}] = BERT([S; <Is Error>; V])}
\end{aligned}
\label{eq_bert}
\end{equation}
,where $l$ is maximum sequence length of the input.

And we get the embedding of start token <s> for each entity as the entity embeddings $HE_{V} = \{h_{ev_1}, h_{ev_2},..., h_{ev_{nv}}\}$ and $HE_{S} = \{h_{es_1}, h_{es_2},..., h_{es_{ns}}\}$ for $V$ and $S$ respectively. We also get $h_{err}$, an embedding of \textit{<Is Error>}.

\paragraph{Error Detection}
Using the computed embeddings, we compute the erroneous score for all of the entities, in summary, using the embedding of \textit{<Is Error>} token $h_{err}$ as follows.
\begin{equation}
\begin{aligned}
\scriptstyle{\hat{s}_{err_{i}} = P(Err|es_i) = sigmoid(h_{es_i}^\intercal~W_{det}h_{err} + b_{det})} 
\end{aligned}
\label{eq_det}
\end{equation}

,where $i=1,2,3, ..., ns$. The $W_{det}$ and $b_{det}$ are model parameters.

\paragraph{Error Correction}
For the entities that are factual errors, we compute the correction score between the entities and all of the entities in the evidence sentences similar to error detection as follows.
\begin{equation}
\begin{aligned}
\scriptstyle{\hat{s}_{cor_{ij}} = P(Cor|es_i, ev_j) = sigmoid(h_{es_i}^\intercal~W_{cor}h_{ev_j} + b_{cor})}
\end{aligned}
\label{eq_cor}
\end{equation}
,where $i=1,2,3, ..., ns_{err}$, $j=1,2,3, ..., nv$. $ns_{err}$ is the number of errors in the summary. The $W_{cor}$ and $b_{cor}$ are model parameters.

\paragraph{Training Objective}
We train the model using binary cross entropy loss for both detection and correction as follows.
\begin{equation}
\begin{aligned}
\scriptstyle{L_{det} = -\frac{\sum_{i=1}^{ns}(s_{err_i}\log({\hat{s}_{err_i}})-(1-s_{err_i})\log({1-\hat{s}_{err_i}}))}{ns}} 
\end{aligned}
\label{eq_det_loss}
\end{equation}

\begin{equation}
\begin{aligned}
\scriptstyle{L_{cor} =  -\frac{\sum_{i=1}^{ns}\sum_{i=1}^{nv}(s_{cor_{ij}}\log({\hat{s}_{cor_{ij}}})-(1-s_{cor_{ij}})\log({1-\hat{s}_{cor_{ij}}}))}{ns\cdot nv}} 
\end{aligned}
\label{eq_cor_loss}
\end{equation}

\begin{equation}
\begin{aligned}
\scriptstyle{L = L_{det} + L_{cor}}
\end{aligned}
\label{eq_all_loss}
\end{equation}
,where ${s}_{err_i}\in\{0,1\}$ and ${s}_{cor_{ij}}\in\{0,1\}$, which are the ground truth labels for detection and correction.

\paragraph{Inference} For the inference stage, we do not have the label as to whether each entity is an error. Therefore, we calculate the two results sequentially, error detection and error correction, using the same BERT embeddings. For each entity, if an erroneous score is above $thr_{det}$, then we let that entity be an error as shown in Figure~\ref{fig_overall}. And then, we search the candidate of correction among the evidence entities $HE_V$, and substitute it with the entity that gets the maximum score as in Figure~\ref{fig_overall}. We conduct correction only when the maximum score is higher than $thr_{cor}$ to prevent unnatural correction caused by failure to find the appropriate entity within the candidate. 

\section{Experiments}
For our experiments, we evaluate our proposed factual error correction method on both synthetic dataset and real-world dataset, based on CNN/DM. We briefly describe the details of two benchmark datasets below.

\subsection{Benchmark Datasets}
Using the same method in Section~\ref{sssec:sec_trainset}, we make a separate 3,000 test tests. As same as the training dataset, the corrupted summaries, and the reference summaries are mixed at the same ratio in this testset. For this synthetic testset, we know the ground truth correction for each summary. 
Hence, we measure the success rate of correction through whether the post-editing model's correction is the same as the ground truth correction.
In addition to this synthetic data, we also use the FactCC-Test set~\cite{kryscinski-etal-2020-evaluating} that has labels on the 503 system-generated summaries whether they are factually consistent or not. Among them, 62 summaries are inconsistent, and 441 summaries are consistent. Different from the synthetic testset, FactCC-Test Dataset does not provide the ground truth correction for the inconsistent summaries. Hence, we manually check the results of all of the systems as in the example of Figure~\ref{fig_case_study}.

\subsection{Implementation Details}
For our experiments, we use \textit{bert-base-cased}\footnote{https://huggingface.co/bert-base-cased} for RFEC. We train the model for five epochs using Adam Optimizer~\cite{kingma2015adam} with a learning rate of 3e-5. For baseline seq2seq model, we use \textit{bart-base}\footnote{https://huggingface.co/facebook/bart-base} following the previous work~\cite{cao2020factual} and train the model using the same dataset we used for training RFEC with same epochs for fair comparison.

\subsection{Performance Comparison}
\paragraph{Synthetic Dataset}
We present the results for the 3k synthetic testset in Table~\ref{table_synthetic}. We observe that the performance of BART is slightly better than RFEC, but our proposed retrieval-based model has a much faster running time. We also observe that accuracy for all of the models is very high for the synthetic dataset since the type of the errors is relatively trivial. 
Also, we find that using only evidence sentences performs slightly lower than using the whole article sentences but have advantages in computing speed for both systems. Especially for RFEC, it does not take much time to calculate the model output, but it costs relatively much time on preprocessing, especially for named entity recognition. And reducing the input length through the sentence selection also reduces the preprocessing time, resulting in faster running time, as shown in Table~\ref{table_synthetic}. For computing the throughput, we make the best effort to set the maximum batch size for each setting using a same environment for a fair comparison. 

\begin{table}[!h]
\centering 
\small
\begin{tabular}{lcc}
\toprule
Method         & Sample/min   & Accuracy \\
\midrule
Seq2seq - BART       & 933 & 90.93 \\
- sentence selection       & 629 & 92.20 \\
RFEC           & 4024 & 91.06 \\
- sentence selection    & 1810 & 91.15 \\
\bottomrule
\end{tabular}
\caption{Factual error correction results on test split of synthetic Test Dataset with the average running time.}
\vspace{-5mm}
\label{table_synthetic}
\end{table}
\paragraph{FactCC-Test Dataset}
We present the results for the FactCC-Test Dataset in Table~\ref{table_k2019}. Compared to the results in the synthetic dataset, both seq2seq and RFEC do not correct many errors, only 9 and 7 for the best settings in both systems among 62 errors. However, as in the synthetic dataset, our proposed method shows almost the same results with eight times less running time compared to the seq2seq method. Also, we can observe that using the correction model also creates a significant number of new errors especially for the seq2seq model without sentence selection. 
\begin{table}[!h]
\centering 
\resizebox{1\columnwidth}{!}{%
\begin{tabular}{lcccc}
\toprule
\multirow{2}{*}{Method} &  \multicolumn{2}{c}{Inconsistent(62)} &  \multicolumn{2}{c}{Consistent(441)} \\ \cline{2-5}
& Changed & Edited & Changed & Edited \\
\midrule
Seq2seq - BART    & 8   &   15  &   2   &   14  \\
- sentence selection   &    9   &   23  &   7   &   78  \\
RFEC        &   7   &   9   &   2   &   23 \\
- sentence selection    &   6   &   8   &   3   &   31  \\
\bottomrule
\end{tabular}
}
\caption{Factual error correction results on FactCC-Testset. Each column represents how many corrections each system has performed for the sample of each label, and how many labels have changed from the correction.
}
\vspace{-5mm}
\label{table_k2019}
\end{table}



\subsection{Qualitative Analysis}
We present the representative success and failure cases of our proposed retrieval-based factual error correction system with the top-3 retrieved entities for the errors in Figure~\ref{fig_case_study}. For the first example, RFEC successfully corrects the error \textit{Valerie Braham} by substituting it with \textit{Philippe Braham} that gets a higher correction score among the entities in the evidence sentences. Also, as the object to be corrected is a person's name, we can observe that other correction candidates are also names. On the other hand, for the second example, although RFEC detects the error \textit{Raymond}, but do not find the correction candidates whose correction score is above $thr_{cor}$. For this example, \textit{Raymond} should be changed to \textit{the front bench}, but the named entity recognition model fails to capture it and leads to missing it from the correction candidate.
\begin{figure}[t]
\small
\begin{framed}
\textbf{Example 1) - Success}\\\

\textbf{\textit{Evidence Sentences:}} Her husband,  \hl{Philippe Braham}, was one of \hl{17} people killed in \hl{January}'s terror attacks in \hl{Paris}. One month after the terror attacks in \hl{Paris}, a gunman attacked a synagogue in \hl{Copenhagen}, \hl{Denmark}, killing \hl{Dan Uzan}, who was working as a security guard for a \hl{bat mitzvah party}.\\

\textbf{\textit{Input Summary:}} 
{\footnotesize
\colorbox[rgb]{0.407504, 0.533890, 0.893678}{\strut Valerie Braham}
\colorbox[rgb]{1, 1, 1}{\strut was one of} 
\colorbox[rgb]{0.807313, 0.846645, 0.964612}{\strut 17}
\colorbox[rgb]{1, 1, 1}{\strut people killed in}
\colorbox[rgb]{0.807313, 0.846645, 0.964612}{\strut January}
\colorbox[rgb]{1, 1, 1}{\strut 's terror attacks in}
\colorbox[rgb]{0.807313, 0.846645, 0.964612}{\strut Paris}
}\\\
\textbf{\textit{Corrected Summary:}} \textit{Philippe Braham} was one of 17 people killed in January's terror attacks in Paris.\\

\textbf{\textit{Top3 Correction Candidates for Valerie Braham:}} \\
Philippe Braham, Dan Uzan, bat mitzvah\\\

\textbf{Example 2) - Failure}\\\

\textbf{\textit{Evidence Sentences:}} \hl{Sawyer Sweeten} grew up before the eyes of millions as a child star on the endearing family sitcom " \hl{Everybody Loves Raymond}." \hl{Sweeten} , best known for his role \hl{Geoffrey Barone} , was visiting family in \hl{Texas}, entertainment industry magazine Hollywood Reporter reported, where he is believed to have shot himself on the front porch. \\

\textbf{\textit{Input Summary:}} 
{\footnotesize
\colorbox[rgb]{1, 1, 1}{\strut He is believed to have shot himself on}
\colorbox[rgb]{0.607504, 0.703890, 0.943678}{\strut Raymond}
}\\\
\textbf{\textit{Corrected Summary:}} He is believed to have shot himself on Raymond.\\

\textbf{\textit{Top3 Correction Candidates for Raymond:}} \\
Everybody Loves Raymond, Geoffrey Barone, Sawyer Sweeten

\end{framed}
\caption{
Case study on our proposed factual error correction system. The entities in the evidence sentences are highlighted. The color on each entity in each input summary represents the erroneous score, and the darker the color, the higher the erroneous score.}
\label{fig_case_study}
\vspace{-5mm}
\end{figure}

\section{Conclusion}
In this paper, we proposed an efficient factual error correction system RFEC based on two retrieval steps. RFEC first retrieves evidence sentences based on textual similarities between the summary and the article for detecting and correcting factual errors. Then, if there is an entity that is a cause of factual errors, RFEC substitutes it with one of the entities in the evidence sentences as a retrieval-based approach. Experiments on two benchmark datasets demonstrate that our proposed method shows competitive results compared to strong baseline seq2seq with a much faster inference speed.
\bibliography{acl}

\begin{thebibliography}{12}
\expandafter\ifx\csname natexlab\endcsname\relax\def\natexlab#1{#1}\fi

\bibitem[{Cao et~al.(2020)Cao, Dong, Wu, and Cheung}]{cao2020factual}
Meng Cao, Yue Dong, Jiapeng Wu, and Jackie Chi~Kit Cheung. 2020.
\newblock Factual error correction for abstractive summarization models.
\newblock In \emph{Proceedings of the 2020 Conference on Empirical Methods in
  Natural Language Processing (EMNLP)}, pages 6251--6258.

\bibitem[{Cao et~al.(2018)Cao, Wei, Li, and Li}]{cao2018faithful}
Ziqiang Cao, Furu Wei, Wenjie Li, and Sujian Li. 2018.
\newblock Faithful to the original: Fact aware neural abstractive
  summarization.
\newblock In \emph{Proceedings of the AAAI Conference on Artificial
  Intelligence}, volume~32.

\bibitem[{Devlin et~al.(2019)Devlin, Chang, Lee, and
  Toutanova}]{devlin2019bert}
Jacob Devlin, Ming-Wei Chang, Kenton Lee, and Kristina Toutanova. 2019.
\newblock \href {https://doi.org/10.18653/v1/N19-1423} {{BERT}: Pre-training of
  deep bidirectional transformers for language understanding}.
\newblock In \emph{Proceedings of the 2019 Conference of the North {A}merican
  Chapter of the Association for Computational Linguistics: Human Language
  Technologies, Volume 1 (Long and Short Papers)}, pages 4171--4186,
  Minneapolis, Minnesota. Association for Computational Linguistics.

\bibitem[{Kingma and Ba(2015)}]{kingma2015adam}
Diederik~P Kingma and Jimmy Ba. 2015.
\newblock Adam: A method for stochastic optimization.
\newblock In \emph{ICLR (Poster)}.

\bibitem[{Kryscinski et~al.(2020)Kryscinski, McCann, Xiong, and
  Socher}]{kryscinski-etal-2020-evaluating}
Wojciech Kryscinski, Bryan McCann, Caiming Xiong, and Richard Socher. 2020.
\newblock \href {https://doi.org/10.18653/v1/2020.emnlp-main.750} {Evaluating
  the factual consistency of abstractive text summarization}.
\newblock In \emph{Proceedings of the 2020 Conference on Empirical Methods in
  Natural Language Processing (EMNLP)}, pages 9332--9346, Online. Association
  for Computational Linguistics.

\bibitem[{Lin(2004)}]{lin-2004-rouge}
Chin-Yew Lin. 2004.
\newblock \href {https://www.aclweb.org/anthology/W04-1013} {{ROUGE}: A package
  for automatic evaluation of summaries}.
\newblock In \emph{Text Summarization Branches Out}, pages 74--81, Barcelona,
  Spain. Association for Computational Linguistics.

\bibitem[{Lin and Ng(2019)}]{lin2019abstractive}
Hui Lin and Vincent Ng. 2019.
\newblock Abstractive summarization: A survey of the state of the art.
\newblock In \emph{Proceedings of the AAAI Conference on Artificial
  Intelligence}, volume~33, pages 9815--9822.

\bibitem[{Maynez et~al.(2020)Maynez, Narayan, Bohnet, and
  McDonald}]{maynez2020faithfulness}
Joshua Maynez, Shashi Narayan, Bernd Bohnet, and Ryan McDonald. 2020.
\newblock On faithfulness and factuality in abstractive summarization.
\newblock In \emph{Proceedings of the 58th Annual Meeting of the Association
  for Computational Linguistics}, pages 1906--1919.

\bibitem[{Nallapati et~al.(2017)Nallapati, Zhai, and
  Zhou}]{nallapati2017summarunner}
Ramesh Nallapati, Feifei Zhai, and Bowen Zhou. 2017.
\newblock Summarunner: A recurrent neural network based sequence model for
  extractive summarization of documents.
\newblock In \emph{Thirty-First AAAI Conference on Artificial Intelligence}.

\bibitem[{Nallapati et~al.(2016)Nallapati, Zhou, dos Santos, Gu̇l{\c{c}}ehre,
  and Xiang}]{nallapati-etal-2016-abstractive}
Ramesh Nallapati, Bowen Zhou, Cicero dos Santos, {\c{C}}a{\u{g}}lar
  Gu̇l{\c{c}}ehre, and Bing Xiang. 2016.
\newblock \href {https://doi.org/10.18653/v1/K16-1028} {Abstractive text
  summarization using sequence-to-sequence {RNN}s and beyond}.
\newblock In \emph{Proceedings of The 20th {SIGNLL} Conference on Computational
  Natural Language Learning}, pages 280--290, Berlin, Germany. Association for
  Computational Linguistics.

\bibitem[{Zhao et~al.(2020)Zhao, Cohen, and Webber}]{zhao-etal-2020-reducing}
Zheng Zhao, Shay~B. Cohen, and Bonnie Webber. 2020.
\newblock \href {https://doi.org/10.18653/v1/2020.findings-emnlp.203} {Reducing
  quantity hallucinations in abstractive summarization}.
\newblock In \emph{Findings of the Association for Computational Linguistics:
  EMNLP 2020}, pages 2237--2249, Online. Association for Computational
  Linguistics.

\bibitem[{Zhu et~al.(2021)Zhu, Hinthorn, Xu, Zeng, Zeng, Huang, and
  Jiang}]{zhu2021enhancing}
Chenguang Zhu, William Hinthorn, Ruochen Xu, Qingkai Zeng, Michael Zeng,
  Xuedong Huang, and Meng Jiang. 2021.
\newblock Enhancing factual consistency of abstractive summarization.
\newblock In \emph{Proceedings of the 2021 Conference of the North American
  Chapter of the Association for Computational Linguistics: Human Language
  Technologies}, pages 718--733.

\end{thebibliography}
\bibliographystyle{acl_natbib}

\appendix
\section{Experimental Details}

\subsection{Reproducibility Checklist}

\paragraph{Computing Infrastructure}
All of the experiments are done using NVIDIA RTX A5000 24G with Python 3.8.8 and PyTorch 1.10.1. We measure the running time, including the preprocessing time of each method using a single A5000 GPU and Intel(R) Xeon(R) Silver 4210R CPU (2.40 GHz). 

\paragraph{Hyperparameters}
We set both $thr_{det}$ and $thr_{cor}$ for 0.5 using the validation set. For maximum sequence length, we set 1024 for BART, 256 for BART without evidence selection, 256 for RFEC, and 512 for RFEC without evidence sentence selection.

\end{document}